\documentclass[11pt]{article}

\usepackage[preprint]{acl}

\usepackage{times}
\usepackage{latexsym}

\usepackage[T1]{fontenc}

\usepackage[utf8]{inputenc}

\usepackage{microtype}

\usepackage{inconsolata}

\usepackage{graphicx}

\usepackage{tikz}
\usepackage{algorithm}
\usepackage{algpseudocode}
\usepackage{algorithmicx}
\usepackage{amsmath}
\usepackage{amssymb}
\usepackage{amsthm}
\usepackage{amsfonts}
\usepackage{booktabs}
\usepackage{multirow}
\usepackage{float}
\usepackage{enumitem}
\usepackage{xspace}
\usepackage{adjustbox}
\usepackage{colortbl}
\usepackage[table]{xcolor}

\newcommand{\mname}{{\sc SAGO}\xspace}

\definecolor{basebg}{HTML}{F5F5F5}     
\definecolor{methodbg}{HTML}{E3F2FD}    
\definecolor{bestcolor}{HTML}{1565C0} 

\definecolor{heat1}{HTML}{F7FBFF}  
\definecolor{heat2}{HTML}{DEEBF7}  
\definecolor{heat3}{HTML}{C6DBEF}  
\definecolor{heat4}{HTML}{9ECAE1}  
\definecolor{heat5}{HTML}{6BAED6}  
\definecolor{heat6}{HTML}{4292C6}  

\title{Modeling LLM Unlearning as an Asymmetric Two-Task Learning Problem}

\author{
    Zeguan Xiao$^{1}$,
    Siqing Li$^{2}$,
    Yong Wang$^{3}$, 
    Xuetao Wei$^{2}$, 
    Jian Yang$^{4}$\\
    \textbf{Yun Chen}$^{1,5}$\thanks{Corresponding Authors.}, 
    \textbf{Guanhua Chen}$^{2}$\footnotemark[1] \\
    $^1$Shanghai University of Finance and Economics,$^3$Alibaba Group\\
    $^2$Southern University of Science and Technology,$^4$Beihang University\\
    $^5$MoE Key Laboratory of Interdisciplinary Research of Computation and Economics
}

\begin{document}
\maketitle
\begin{abstract}
Machine unlearning for large language models (LLMs) aims to remove targeted knowledge while preserving general capability. In this paper, we recast LLM unlearning as an asymmetric two-task problem: retention is the primary objective and forgetting is an auxiliary. From this perspective, we propose a retention-prioritized gradient synthesis framework that decouples task-specific gradient extraction from conflict-aware combination. Instantiating the framework, we adapt established PCGrad to resolve gradient conflicts, and introduce \mname, a novel retention-prioritized gradient synthesis method. Theoretically, both variants ensure non-negative cosine similarity with the retain gradient, while \mname achieves strictly tighter alignment through constructive sign-constrained synthesis. Empirically, on WMDP Bio/Cyber and RWKU benchmarks, \mname consistently pushes the Pareto frontier: e.g., on WMDP Bio (SimNPO+GD), recovery of target model MMLU performance progresses from 44.6\% (naive) to 94.0\% (+PCGrad) and further to 96.0\% (+\mname), while maintaining comparable forgetting strength. Our results show that re-shaping gradient geometry, rather than re-balancing losses, is the key to mitigating unlearning-retention trade-offs. 
\end{abstract}

\section{Introduction}

\begin{figure*}
\centering
\includegraphics[width=0.9\linewidth]{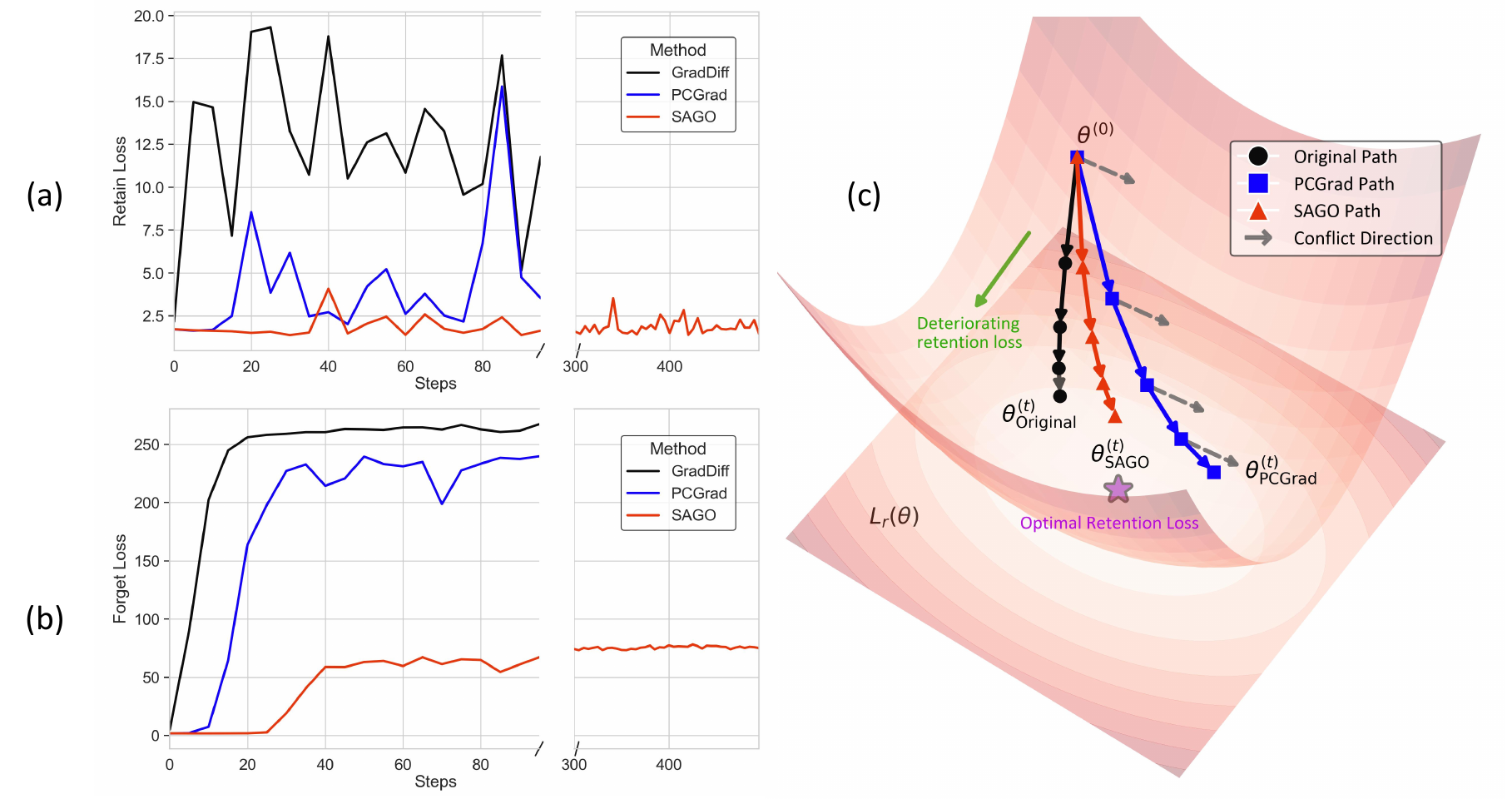}
\caption{\textbf{Visualization of loss dynamics in LLM unlearning and retention-prioritized frameworks.} Panels (a) and (b) show retain and forget losses on the WMDP Biosecurity benchmark using GradDiff, PCGrad, and SAGO. SAGO outperforms in maintaining low retain loss while achieving high forget performance, indicating reduced gradient conflicts and improved retention. Panel (c) shows that while GradDiff (Original) struggles with retention, PCGrad and SAGO dynamically refine gradients, achieving effective unlearning with stable retention.}
\label{fig:main}
\end{figure*}

Large language models (LLMs) have achieved remarkable success in recent years. However, like many powerful technologies, LLMs are inherently dual-use and can be leveraged for both beneficial and harmful purposes. LLMs are trained on vast corpora collected from the Internet, which unavoidably contain personal information and potentially hazardous knowledge. Their capacity to memorize and reproduce training data can therefore be exploited to disclose sensitive information or to generate harmful content. A common mitigation is alignment training, which aims to teach LLMs to refuse harmful queries. Nevertheless, recent studies \citep{zou2023universal,yuan2023cipherchat,xiao-etal-2024-distract} find that adversaries can easily craft jailbreak prompts to circumvent these safeguards.

To address these vulnerabilities, machine unlearning (MU) \citep{cao2015towards} has emerged as a promising solution to mitigate the risks associated with LLMs by directly removing private information and hazardous knowledge from the model. Unlearned models offer stronger inherent safety because even if they are jailbroken, they lack the knowledge necessary to enable malicious users.
However, LLM unlearning faces a central challenge: The unlearning often degrades the model's performance, leading to a trade-off between effective unlearning and preserving essential capabilities \citep{wang2025gru}. 

To make the above challenge concrete, we begin with the canonical method commonly used in unlearning: gradient ascent (GA) on the forget set. This method, while simple and directly enforcing forgetting, often leads to over-forgetting and significant performance degradation.
To mitigate this, methods such as NPO \citep{zhang2024negative} and SimNPO \citep{fan2024simplicity} regularize GA in two ways: (i) they transform the unbounded GA objective into a bounded one, which helps prevent catastrophic collapse; and (ii) they apply adaptive smoothing to the forget-set gradients, enabling more controlled divergence during unlearning.
Another line of work, gradient difference (GradDiff \citep{liu2022continual}), couples GA on the forget set with gradient descent (GD) on a retain set to preserve core capabilities.
Despite these advances, conflicts between forget and retain gradients persist.

To address conflicts between forgetting and retaining gradients, \citet{reisizadeh2025blur} recently proposed a bi-level optimization approach for LLM unlearning that prioritizes the forgetting objective over the retaining one.
\textbf{In this work}, we model the trade-off between unlearning and retention as an asymmetric two-task learning problem, with retention as the primary task and unlearning as the auxiliary task. 
We explore two approaches to synthesize gradients. First, we adapt PCGrad \citep{yu2020gradient}, a technique originally designed for mitigating gradient conflicts in multi-task learning to the unlearning scenario. This adaptation ensures that the gradients driving the unlearning process do not conflict destructively with those preserving the model's utility. Second, we propose \mname, a novel retention-prioritized gradient synthesis method that enhances unlearning efficacy without compromising retention performance. The key insight of \mname lies in enforcing element-wise sign alignment between the synthesized gradients and the retention gradients, ensuring the update direction consistently supports retention. An intuitive visualization of our retention-prioritized framework can be found in Figure \ref{fig:main} (c).
We conduct experiments on two widely used LLM unlearning benchmarks, WMDP \citep{li2024wmdp} and RWKU \citep{jin2024rwku}, and demonstrate that both PCGrad and \mname significantly improve retention performance while maintaining competitive unlearning effectiveness compared to vanilla unlearning objectives.
As shown in Figure \ref{fig:main} (a) and (b), our proposed \mname method demonstrates particularly strong performance, notably achieving superior retention with effective unlearning.

Our contributions are summarized as follows\footnote{Code: \url{https://github.com/sustech-nlp/SAGO}}:
\begin{itemize}[leftmargin=*,label=$\triangleright$]
    \item \textbf{Asymmetric formulation.} We reframe LLM unlearning as an asymmetric two-task problem and show that viewing retention as the primary objective leads to an effective generic framework. The resulting framework integrates seamlessly with diverse unlearning objectives, including existing GA+GD, NPO+GD, and SimNPO+GD, and is readily extensible to future objectives.
    \item \textbf{New gradient synthesis methods.} We adapt the established PCGrad to resolve gradient conflicts, and introduce \mname, a novel retention-prioritized gradient synthesis method. Theoretically, both variants ensure non-negative cosine similarity with the retain gradient, while \mname achieves strictly tighter alignment through element-wise sign-constrained synthesis.
    \item \textbf{Empirical gains.} \mname consistently improves retention at comparable forgetting. On WMDP, MMLU gains are 17.8--30.7 points (Bio) and 4.1--11.7 points (Cyber) over the naive method, with an additional 0.4--1.2 points over PCGrad, keeping comparable or better forgetting effectiveness. Similar improvements are observed on RWKU.
\end{itemize}

\section{Preliminaries}

\subsection{Problem Formulation}

Given an original model $\mathcal{M}$ that is already trained on a dataset $\mathcal{D}$, Machine Unlearning (MU) \citep{cao2015towards} aims to remove specific information from $\mathcal{M}$, resulting in an unlearned model $\mathcal{M}'$ that no longer retains or utilizes this undesired information.
Formally, we define the information to forget as a subset of $\mathcal{D}$, called the \textit{forget set} $\mathcal{D}_f$.
Ideally, after unlearning, the model should behave as if trained on \textit{retain set} \( \mathcal{D}_r = \mathcal{D} \setminus \mathcal{D}_f\).

In the context of LLM unlearning, the forget set $\mathcal{D}_f$ and retain set $\mathcal{D}_r$ are typically text corpora. The unlearning process involves finetuning the original model $\mathcal{M}$ on $\mathcal{D}_f$ and/or $\mathcal{D}_r$ with specific objectives to obtain $\mathcal{M}'$.

\subsection{LLM Unlearning Methods}

We denote the probability distribution defined by an LLM with parameters \(\boldsymbol{\theta}\) as \(p(x; \boldsymbol{\theta})\), where \(x\) represents a text sequence.

The standard unlearning objective is to suppress the model's likelihood on the forget set $\mathcal{D}_f$—that is, drive $\log p(x;\boldsymbol{\theta})$ downward for $x\in\mathcal{D}_f$. 
This is implemented by performing gradient ascent (GA) on the cross-entropy objective (equivalently, minimizing the negative cross-entropy) over $\mathcal{D}_f$:
\begin{equation*}
\mathcal{L}_{\mathrm{GA}}(\mathcal{D}_f; \boldsymbol{\theta})
= - \mathbb{E}_{x \sim \mathcal{D}_f}\big[-\log p(x; \boldsymbol{\theta})\big].
\end{equation*}
Minimizing the above GA objective reduces the assigned probabilities $p(x;\boldsymbol{\theta})$, achieving the goal of minimizing the forget-set likelihood.

Given the unbounded nature of GA, it can lead to over-forgetting and significant performance degradation. To mitigate this, methods such as NPO \citep{zhang2024negative} and SimNPO \citep{fan2024simplicity} regularize GA by transforming the unbounded objective into a bounded one and applying adaptive smoothing to the forget-set gradients. This allows for more controlled divergence during unlearning, preventing catastrophic collapse. Formally, their objectives can be written as:
\begin{equation*}
\resizebox{\columnwidth}{!}{$
\mathcal{L}_{\mathrm{NPO}}(\boldsymbol{\theta})
= -\frac{2}{\beta}\,\mathbb{E}_{x\sim \mathcal{D}_f}
\log \sigma\!\Big(
-\beta \log \frac{p(x; \boldsymbol{\theta})}{p(x; \boldsymbol{\theta}_{\mathrm{ref}})}
\Big),
$}
\end{equation*}
\begin{equation*}
\resizebox{\columnwidth}{!}{$
\mathcal{L}_{\mathrm{SimNPO}}(\boldsymbol{\theta})
= -\frac{2}{\beta}\,\mathbb{E}_{x\sim \mathcal{D}_f}
\log \sigma\!\Big(
-\frac{\beta}{|x|}\,\log p(x; \boldsymbol{\theta}) - \gamma
\Big).
$}
\end{equation*}
Here, $p(x; \boldsymbol{\theta}_{\mathrm{ref}})$ is the probability distribution of pre-unlearning model, $\sigma(\cdot)$ is the logistic sigmoid, $\beta>0$ controls the sharpness (smoothing) of the bounded transformation, $|x|$ is the length of text sequence, and $\gamma$ is an margin to further suppress the likelihood of forget set.

A common practice to preserve the model's core capabilities during unlearning is to incorporate a retain objective on the retain set $\mathcal{D}_r$:
\begin{equation*}
\mathcal{L}_{\mathrm{GD}}(\mathcal{D}_r; \boldsymbol{\theta})
= \mathbb{E}_{x \sim \mathcal{D}_r}\big[-\log p(x; \boldsymbol{\theta})\big].
\end{equation*}

Building upon the above components, we write a generic unlearning objective as $\mathcal{L}_{unlearn}$:
\begin{equation}
\label{eq:final_unlearning_objective}
\mathcal{L}_{\mathrm{unlearn}}(\boldsymbol{\theta})
= \gamma \,\mathcal{L}_{f}(\mathcal{D}_f; \boldsymbol{\theta})
+ \alpha \,\mathcal{L}_{\mathrm{GD}}(\mathcal{D}_r; \boldsymbol{\theta}),
\end{equation}
where $\mathcal{L}_{f}$ can be instantiated by $\mathcal{L}_{\mathrm{GA}}$, $\mathcal{L}_{\mathrm{NPO}}$, or $\mathcal{L}_{\mathrm{SimNPO}}$. $\gamma $ and $\alpha$ are hyperparameters balancing the two objectives.

\paragraph{GradDiff as a Special Case.}
The classical Gradient Difference (GradDiff) couples GA on the forget set with GD on the retain set by choosing $\mathcal{L}_{f}=\mathcal{L}_{\mathrm{GA}}$ in Eq.~\ref{eq:final_unlearning_objective}, yielding:
\begin{equation*}
\label{eq:graddiff}
\mathcal{L}_{\mathrm{GradDiff}}(\boldsymbol{\theta})
= \gamma \,\mathcal{L}_{\mathrm{GA}}(\mathcal{D}_f; \boldsymbol{\theta})
+ \alpha \,\mathcal{L}_{\mathrm{GD}}(\mathcal{D}_r; \boldsymbol{\theta}).
\end{equation*}
Replacing $\mathcal{L}_{\mathrm{GA}}$ by $\mathcal{L}_{\mathrm{NPO}}$ or $\mathcal{L}_{\mathrm{SimNPO}}$ yields the corresponding NPO+GD and SimNPO+GD variants under the unified objective Eq.~\ref{eq:final_unlearning_objective}.

\section{Methodology} \label{sec:methodology}

\subsection{Motivation} \label{sec:motivation}

The generic unlearning objective in Eq.~\ref{eq:final_unlearning_objective} shows that LLM unlearning is a \emph{two-task learning} problem: one task drives the model to \emph{forget}, while the other task \emph{retains} the general ability learned from the retain set. At first sight, this looks similar to standard multi-task learning (MTL). However, unlearning has a fundamental \textbf{asymmetry}: retention is the \emph{primary} objective and forgetting is an \emph{auxiliary} objective applied under a \textit{do-no-harm} constraint. We do not seek a balanced compromise between tasks; instead, we wish to \textbf{(i)} preserve performance on the retain set and \textbf{(ii)} remove specific information with minimal side effects. This asymmetric preference makes many MTL methods, whose goal is to equalize task progress or fairness, suboptimal or even harmful \citep{chen2020just,liu2021conflict,pmlr-v162-navon22a}.

The specificity of the unlearning problem motivates a shift from \textit{loss balancing} to \textbf{retention-prioritized gradient synthesis}. Our perspective is to treat the retain gradient as the anchor direction and inject forgetting only where it does not fight retention. Our methods are inspired by this principle, as detailed next.

\begin{algorithm*}[ht]
\caption{Framework of \mname.}
\label{alg:sago}
\begin{algorithmic}[1]
\Require Initial parameters $\theta$, Forget set $\mathcal{D}_f$, Retain set $\mathcal{D}_r$, Number of iterations $T$, Learning rate $\eta$
\State Initialize $\theta^0 \leftarrow \theta$
\For{$t \gets 1$ \textbf{to} $T$}
    \State Sample batch $B_f \sim \mathcal{D}_f$ and $B_r \sim \mathcal{D}_r$
    \State $g_f^t \gets \nabla_{\theta^{t-1}} \mathcal{L}_{f}(B_f; \theta^{t-1})$ \Comment{Gradient on forget set}
    \State $g_r^t \gets \nabla_{\theta^{t-1}} \mathcal{L}_{r}(B_r; \theta^{t-1})$ \Comment{Gradient on retain set}
    
    \State $g_{\text{final}}^t \gets \Call{CombineGradients}{g_r^t, g_f^t}$ \Comment{Use PCGrad or SAGO}
    \State $\theta^t \gets \theta^{t-1} - \eta \cdot g_{\text{final}}^t$ \Comment{Update model parameters}
\EndFor
\State \Return Unlearned Model $\mathcal{M}'$ with parameters $\theta^T$
\end{algorithmic}
\end{algorithm*}

\subsection{Framework Overview}

Our unlearning procedure (Algorithm~\ref{alg:sago}) operates as a two-stage iterative optimization that alternates between (i) extracting task-specific gradients and (ii) synthesizing a conflict-aware update direction. Each iteration (Lines~3-5) draws mini-batches from the forget set $\mathcal{D}_f$ and retain set $\mathcal{D}_r$ and computes their respective gradients $g_f^t=\nabla_{\theta^{t-1}}\mathcal{L}_{f}(B_f;\theta^{t-1})$ and $g_r^t=\nabla_{\theta^{t-1}}\mathcal{L}_{r}(B_r;\theta^{t-1})$. Line~6 encapsulates the core design choice: \textsc{CombineGradients} produces a final update direction $g_{\text{final}}^t$ that injects forgetting gradients only to the extent that they do not harm retention.

Crucially, Algorithm~\ref{alg:sago} treats gradient synthesis as a modular component: different conflict-mitigation methods can be plugged into \textsc{CombineGradients}. In this work, we explore two methods: (i) Project Conflicting Gradients (PCGrad) \citep{yu2020gradient} and (ii) our proposed novel Sign-Align Gradient Optimization (\mname). Their mechanisms and theoretical properties are detailed in the subsequent subsections.

\subsection{Project Conflicting Gradients (PCGrad)}

In multi-task learning, conflicting gradients between tasks can hinder optimization and degrade performance. To address this, PCGrad was proposed \cite{yu2020gradient}, which resolves conflicts by projecting a task's gradient onto the normal plane of another task's gradient when their directions conflict. Motivated by the discussion in Section \ref{sec:motivation}, we project the forget gradient to prevent it from interfering with the retain gradient if they conflict (i.e. $g_f^{\top} g_r < 0$), thereby prioritizing retention:
\begin{equation*}
\tilde g_f = g_f - \frac{g_f \cdot g_r}{||g_r||^2} \cdot g_r,
\end{equation*}
where $\frac{g_f \cdot g_r}{||g_r||^2}$ computes the projection of the entire gradient vector $g_f$ onto $g_r$. 

The official PCGrad \cite{yu2020gradient} flattens all parameters into a single vector and performs the projection in this joint space. GRU \cite{wang2025gru}, a recent unlearning method, follows the same approach.
We instead apply a module-wise projection. For each module $j$, let $g_f^{j}$ and $g_r^{j}$ denote the gradients of the forget and retain objectives with respect to its parameter vector $\theta_j$. We detect conflict locally and only then modify the forget gradient:
\begin{equation*}
\tilde g_f^{j} = g_f^{j} - \frac{g_f^{j} \cdot g_r^{j}}{\lVert g_r^{j} \rVert^2}\; g_r^{j}.
\label{eq:projection}
\end{equation*}
This localized projection: (i) prevents conflicts in one module from triggering unnecessary correction elsewhere, (ii) yields finer-grained mitigation that empirically enhances retention performance \citep{liu2025modular}.
The final gradient is then synthesized as a weighted combination of the retain gradient and the modified forget gradient:
\begin{equation*}
g_{\mathrm{final}}^{j} \;=\; \alpha\, g_r^{j} + \gamma\, \tilde g_f^{j}.
\label{eq:PCGrad}
\end{equation*}


\subsection{Sign-Align Gradient Optimization (SAGO)}

The core idea of \mname is to construct an update direction that not only effectively removes information in the forget set but also minimizes disruption to general knowledge as much as possible.
A central challenge is that gradients are inherently noisy. For example, the gradient on the forget task may embed components related to general linguistic competence or general-domain knowledge. Naively combining forget and retain gradients, therefore, often induces degradation in retention performance.
While PCGrad mitigates part of this issue by projecting the forget gradient onto the orthogonal complement of the retain gradient, it can still be suboptimal: the retain gradient itself is an imperfect estimator, and the projection may offer limited protection against performance degradation.

Motivated by empirical findings that different parameters specialize in distinct functions \citep{geva2021transformer,meng2022locating}, we posit that forgetting and retention signals need not act uniformly across all weights. Accordingly, \mname applies a fine-grained, per-parameter (element-wise) gradient synthesis to inject forgetting only where it does not conflict with retention.

Concretely, we treat parameters whose forget and retain gradients have opposite signs as carriers of general knowledge: in those dimensions, the ``un-forget'' direction (the negative of the forget gradient) and the retain direction are aligned, suggesting the retain gradient should be preserved while suppressing the contribution of the forget gradient. Conversely, when the signs match, we regard the dimensions as task-specific and free of conflict, and we allow the forget gradient to pass through.

Formally, \mname first gates the two task gradients element-wise:
\begin{equation*}
\tilde g_f = g_f \odot \mathbb{I}(g_f \odot g_r \ge 0),
\end{equation*}
\begin{equation*}
\tilde g_r = g_r \odot \mathbb{I}(g_f \odot g_r < 0),
\end{equation*}
where $\odot$ denotes element-wise multiplication, and $\mathbb{I}(\cdot)$ is the indicator function (1 if the condition holds, 0 otherwise).

The final gradient is then synthesized as a weighted combination of the gated forget and retain gradients:
\begin{equation*}
g_{\text{final}} = \alpha\,\tilde g_r + \gamma\,\tilde g_f.
\end{equation*}

\mname yields two coupled effects that are central to its retention-prioritized behavior.
First, $\tilde g_f$ and $\tilde g_r$ are orthogonal by construction: they have disjoint support, so $ \tilde g_f^{\top} \tilde g_r = 0$. This orthogonality eliminates direct conflicts between the two tasks.
Second, the final update direction remains strictly aligned with the retain gradient, and no coordinate in the final update ever points against the retain signal; therefore, the step preserves the coarse directional geometry of the retention objective while still injecting forgetting pressure where it is provably non-harmful.
ith $g_r$ compared to PCGrad. The analysis assumes vector gradients and equal weights $\alpha = \gamma = 1$.

\begin{figure}[t]
    \centering
    \includegraphics[width=0.9\linewidth]{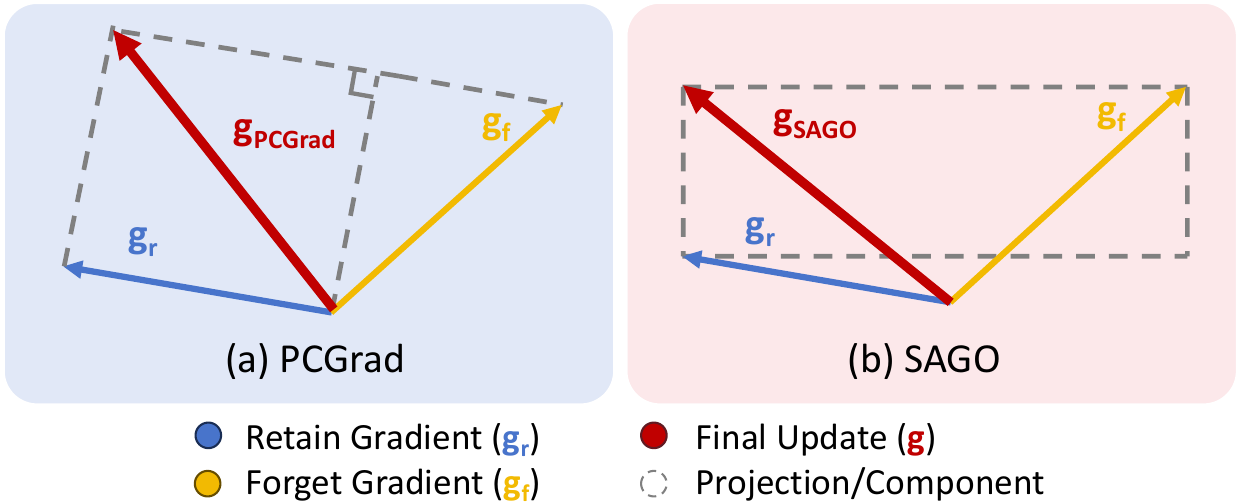}
    \caption{\textbf{Illustration of final update gradients (red) in PCGrad (a) and SAGO (b).} For PCGrad, the forget gradient ($g_f$) is projected orthogonally onto the retain gradient ($g_r$), and the resulting projected vector is then combined with $g_r$. For SAGO, when the two gradients conflict, $g_r$ is used, and when the gradients align, $g_f$ is applied. The updates produced by SAGO demonstrate a higher degree of alignment with $g_r$}
    \label{fig:gradient_alignment}
\end{figure}

\subsection{Theoretical Analysis} \label{sec:theoretical_analysis}

In LLM unlearning, preserving general knowledge requires the final update direction to align closely with the retain gradient ($g_r$), minimizing disruption to existing knowledge. We demonstrate that both PCGrad and \mname ensure non-negative cosine similarity between their final gradients and $g_r$, confirming acute angular alignment. Furthermore, we prove that SAGO achieves superior alignment under equal weighting ($\alpha = \gamma = 1$).

Recall the cosine similarity definition: $\cos \theta = \frac{g_{\mathrm{final}}^\top g_r}{\|g_{\mathrm{final}}\|\|g_r\|}$. For PCGrad, orthogonal projection ensures $\tilde{g}_f^{\text{PCGrad}} \perp g_r$, simplifying the dot product:
\begin{align*}
g_{\text{final}}^{\text{PCGrad}} \cdot g_r &= g_r \cdot g_r + \tilde{g}_f^{\text{PCGrad}} \cdot g_r \\
&= \|g_r\|^2.
\end{align*}
The cosine similarity then becomes:
\[
\cos \theta_{\mathrm{P}} = \frac{\|g_r\|^2}{\|g_{\mathrm{final}}^{\text{PCGrad}}\| \|g_r\|} = \left(1 + \frac{\|\tilde{g}_f\|^2}{\|g_r\|^2}\right)^{-1/2},
\]
guaranteeing $\cos \theta_{\mathrm{P}} \geq 0$.

SAGO employs gradient gating with disjoint supports: $\tilde{g}_f$ operates solely on aligned dimensions $S = \{i : g_{f}^i g_{r}^i \geq 0\}$, while $\tilde{g}_r$ operates on conflicting dimensions $C = \{i : g_{f}^i g_{r}^i < 0\}$. This yields $\tilde{g}_f \perp \tilde{g}_r$ and produces:
\begin{equation*}
\cos \theta_{\mathrm{S}} = \frac{\sum_{i \in C} (g_{r}^i)^2 + \sum_{i \in S} g_{f}^i g_{r}^i}{\|g_{\mathrm{final}}^{\text{SAGO}}\| \|g_r\|}.
\end{equation*}
Sign alignment in $S$ ensures $\sum_{i \in S} g_{f}^i g_{r}^i > 0$, yielding $\cos \theta_{\mathrm{S}} > 0$.

\begin{table*}[ht]
\centering
\resizebox{0.94\textwidth}{!}{
\begin{tabular}{lccccccccccc}
\toprule
\multirow{3}{*}{\textbf{Method}} 
 & \multicolumn{4}{c}{\textbf{WMDP}} 
 & \multicolumn{6}{c}{\textbf{RWKU}} \\
\cmidrule(lr){2-5}\cmidrule(lr){6-11}
 & \multicolumn{2}{c}{\textbf{Bio}} & \multicolumn{2}{c}{\textbf{Cyber}} 
 & \multicolumn{3}{c}{\textbf{Forget Set $\downarrow$}} 
 & \multicolumn{3}{c}{\textbf{Neighbor Set $\uparrow$}} \\
\cmidrule(lr){2-3}\cmidrule(lr){4-5}\cmidrule(lr){6-8}\cmidrule(lr){9-11}
 & \textbf{Forget $\downarrow$} & \textbf{MMLU $\uparrow$} & \textbf{Forget $\downarrow$} & \textbf{MMLU $\uparrow$} & \textbf{FB} & \textbf{QA} & \textbf{All} & \textbf{FB} & \textbf{QA} & \textbf{All} \\
\midrule
\rowcolor{basebg} Target Model & 64.4 & 59.8 & 44.4 & 59.8 & 75.2 & 91.1 & 83.2 & 78.0 & 92.1 & 85.1 \\
\midrule
GA        & 24.7 & 24.7 & 23.4 & 26.6 & 3.1 & 3.1 & 3.1 & 5.4 & 6.0 & 5.7 \\
NPO       & 27.1 & 30.8 & 34.7 & 52.9 & 8.3 & 3.1 & 5.7 & 10.2 & 6.2 & 8.2 \\
SimNPO    & 24.8 & 27.1 & 31.4 & 40.0 & 9.2 & 4.3 & 6.8 & 12.5 & 7.9 & 10.2 \\
RMU       & 28.0 & 50.1 & 27.7 & 57.8 & - & - & - & - & - & - \\
\midrule
GA + GD        & 24.7 & 25.1 & 24.7 & 55.6 & 14.1 & 7.3 & 10.7 & 17.4 & 13.5 & 15.5 \\
\rowcolor{methodbg} \quad + PCGrad            & \textbf{24.5} & 53.0 & 27.0 & 58.5 & \textbf{3.1} & \textbf{1.5} & \textbf{2.3} & 13.7 & 23.8 & 18.8 \\
\rowcolor{methodbg} \quad + \mname            & 26.0 & \textbf{54.1} & \textbf{26.2} & \textbf{59.7} & 4.9 & 2.8 & 3.9 & \textbf{24.4} & \textbf{39.8} & \textbf{32.1} \\
\midrule
NPO + GD                  & 30.5 & 38.2 & 31.8 & 46.8 & 10.0 & 7.8 & 8.9 & 13.9 & 12.2 & 13.1 \\
\rowcolor{methodbg} \quad + PCGrad            & 32.9 & 55.4 & 30.8 & 57.5 & \textbf{3.6} & \textbf{2.3} & \textbf{3.0} & 29.9 & 26.8 & 28.4 \\
\rowcolor{methodbg} \quad + \mname            & \textbf{30.0} & \textbf{56.0} & \textbf{29.6} & \textbf{58.5} & 5.1 & 5.7 & 5.4 & \textbf{36.1} & \textbf{47.2} & \textbf{41.7} \\
\midrule
SimNPO + GD               & 26.1 & 26.7 & 31.4 & 51.4 & 10.6 & 7.3 & 9.0 & 44.4 & 45.4 & 44.9 \\
\rowcolor{methodbg} \quad + PCGrad            & 28.7 & 56.4 & 31.1 & 57.9 & 12.7 & 13.9 & 13.3 & 48.8 & 58.1 & 53.5 \\
\rowcolor{methodbg} \quad + \mname            & \textbf{28.2} & \textbf{57.4} & \textbf{29.1} & \textbf{58.3} & \textbf{12.5} & \textbf{13.3} & \textbf{12.9} & \textbf{50.1} & \textbf{63.7} & \textbf{56.9} \\
\bottomrule
\end{tabular}
}
\caption{\textbf{Experimental results on WMDP and RWKU benchmarks.} For WMDP, lower forget performance (accuracy) is better, while higher MMLU (accuracy) reflects better retention. For RWKU, lower ROUGE-L on the Forget Set is better and higher ROUGE-L on the Neighbor Set reflects better retention. The top-performing results in each combination group are highlighted in bold to ease reference.}
\label{tab:main_results}
\end{table*}

As illustrated in Figure \ref{fig:gradient_alignment}, SAGO demonstrates a stronger alignment with $g_r$ compared to PCGrad. This advantage can be attributed to two key mechanisms. First, the projection operation in PCGrad can generate antagonistic components when $|\tilde{g}_f^i| > |g_{r}^i|$ in a particular dimension $i$, with $\tilde{g}_f^i$ dominating the final update direction in this dimension. This would cause the sign of the final update direction to be opposite to the original retain gradient, thereby decreasing the value of $g_{\mathrm{final}} ^{\top} g_r$. In contrast, SAGO completely avoids such detrimental opposition by ensuring ${g_{\mathrm{final}}}^i g_{r}^i \geq 0$ for all $i$. Additionally, SAGO employs element-wise gating, enabling fine-grained suppression of over-correction and better preservation of magnitude ratios. In contrast, the unified projection of PCGrad lacks such precise adjustment capabilities. Therefore, SAGO achieves superior directional fidelity with $g_r$, leading to a retention-prioritized gradient.

\begin{figure*}[ht]
    \centering
    \includegraphics[width=0.96\textwidth]{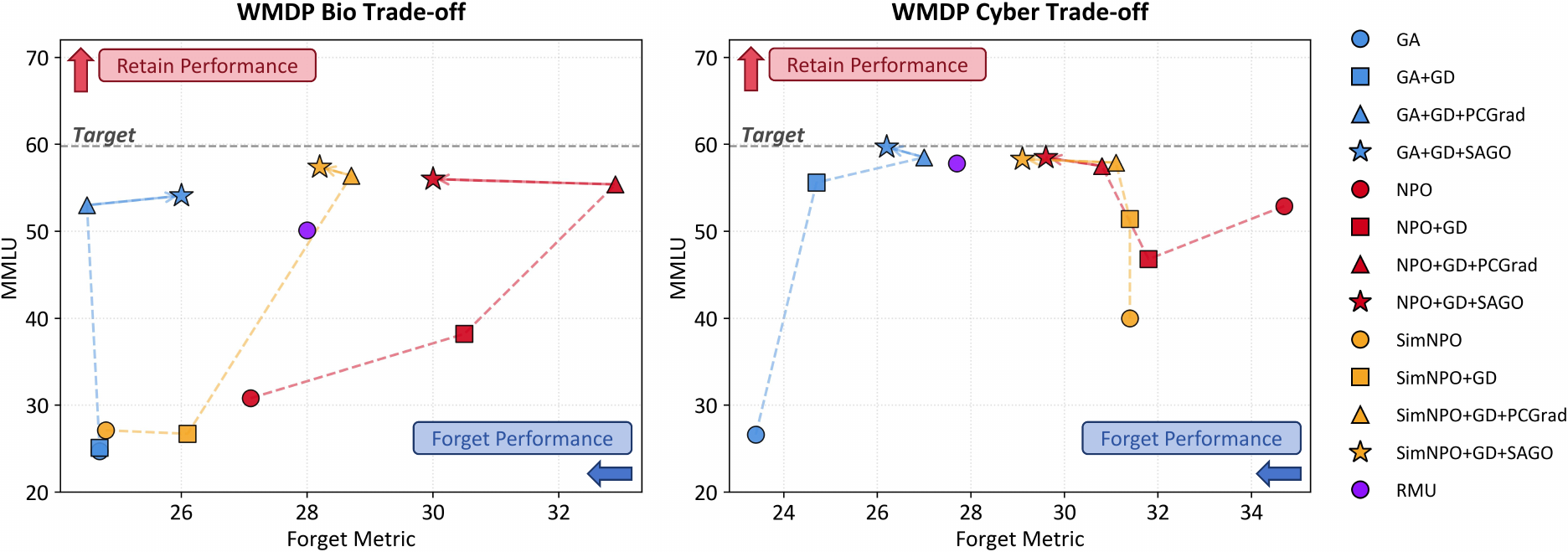}
    \caption{\textbf{Performance comparison across different methods on WMDP benchmark.} 
    The plot provides a visualization of the trade-offs between forget and retain performance on WMDP  Bio (left) and Cyber (right). A smaller forget metric indicates better forgetting effectiveness, while a larger MMLU value reflects better retention performance. Dashed lines connect base methods to their enhanced variants within the same family (same color). The horizontal grey dashed line represents the original model's performance (Target). The reported results highlight that \mname (stars) variants consistently push the Pareto frontier upward for comparable forgetting effectiveness.}
    \label{fig:wmdp_tradeoff}
\end{figure*}

\begin{figure}[ht]
    \centering
    \includegraphics[width=0.48\textwidth]{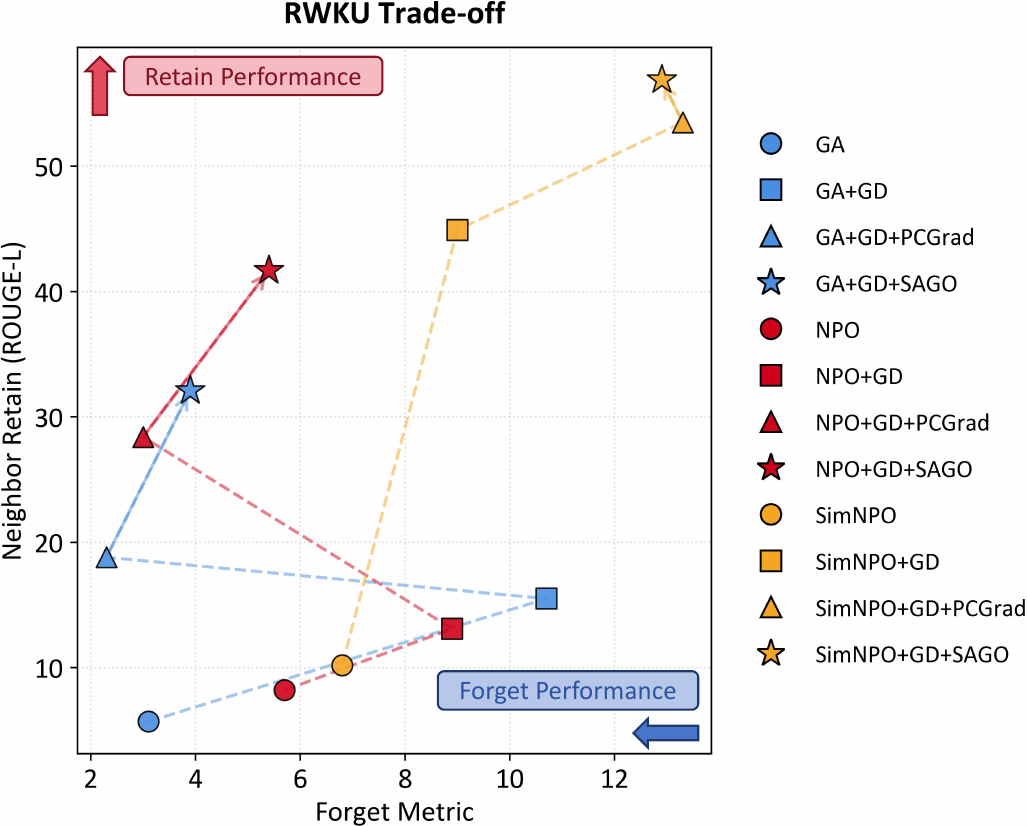}
    \caption{\textbf{Performance comparison across different methods on RWKU benchmark.} The plot provides a visualization of the trade-offs between forget and retain performance on RWKU. 
    A smaller forgetting ROUGE-L indicates better forgetting effectiveness, while a larger retention ROUGE-L reflects better retention performance. 
    The reported results highlight that \mname produces a better frontier than baselines and PCGrad.}
    \label{fig:rwku_tradeoff}
\end{figure}

\section{Experiments}

\subsection{Setup}
\paragraph{Benchmarks.}
We conduct experiments on two widely used LLM unlearning benchmarks: WMDP \citep{li2024wmdp} and RWKU \citep{jin2024rwku}.
WMDP contains expert-written multiple-choice questions in biosecurity, cybersecurity, and chemistry domains. Following \citet{li2024wmdp}, we use the provided forget corpus and use Wikitext \citep{merity2016pointer} as the retain set. We focus on biosecurity and cybersecurity domains, since the forget corpus for the chemistry domain is not publicly available.
RWKU includes 200 real-world famous people as the unlearning targets and provides a forget corpus for each target. We adapt the challenging batch-unlearning setting \citep{jin2024rwku}, in which multiple targets are forgotten simultaneously. From the 200 targets, we select 50 targets as the forget targets and use their corresponding Wikipedia passages as the forget corpus. Since RWKU does not provide a retain set, we construct one using the Wikipedia passages for the 50 remaining targets.

\paragraph{Models.}
Following \citet{li2024wmdp}, for WMDP, we use Zephyr-7B-beta \citep{tunstall2023zephyr} as the target model. For RWKU, we use LLaMA3-8B-Instruct \citep{dubey2024llama}, which is the same as \citet{jin2024rwku}.

\paragraph{Baselines.}
The baselines include two main categories. The first category includes methods that only use a forget objective. This includes GA, NPO \citep{zhang2024negative}, and SimNPO \citep{fan2024simplicity}.
The second category combines a forget objective with a retain objective (i.e., gradient descent on the retain corpus, denoted as GD). This includes GradDiff \citep{liu2022continual} (equivalent to GA + GD), NPO + GD, and SimNPO + GD. For this category of baselines, we evaluate the effectiveness of PCGrad and \mname to resolve conflicts between the forget and retain tasks.
For WMDP, we also include the RMU \citep{li2024wmdp} as a baseline, which is proposed alongside the WMDP benchmark.

\paragraph{Evaluation.}
Following \citet{li2024wmdp}, for WMDP, we report forget effectiveness as accuracy on the benchmark's multiple-choice questions, and retention as accuracy on MMLU \citep{hendryckstest2021mmlu}.
For RWKU, we use fill-in-the-blank (FB) and question-answer (QA) style probes to evaluate the model's ability to recall knowledge and apply it to downstream tasks. Following \citet{jin2024rwku}, we prompt the unlearned model to answer these probes and use ROUGE-L recall score to measure the similarity between the model's predictions and the ground truth answers. The forget performance is evaluated on the 50 forget targets, and the retention is measured on the holdout neighbor targets of these forget targets.


\subsection{Experimental Results}

We present the full quantitative results in Tables~\ref{tab:main_results}, and highlight three observations.

\paragraph{(1) Retention-prioritized gradient synthesis markedly improves the trade-off.} Across many settings, integrating a retain objective (``+GD'') already improves retention over pure forgetting (GA / NPO / SimNPO). However, retention-prioritized synthesis is decisive: replacing naive summation with PCGrad yields large gains, and \mname further improves retention while preserving competitive forgetting. For instance, on WMDP Bio with SimNPO+GD, MMLU rises from 26.7 (naive) to 56.4 (+PCGrad) and 57.4 (+\mname), recovering 96.0\% of the target model performance (59.8) while maintaining low forget accuracy (28.2 vs 26.1 baseline). Similar patterns hold in Cyber and RWKU as well.

\paragraph{(2) \mname consistently matches or exceeds PCGrad on retention with minimal sacrifice in forgetting.} On WMDP Bio, SAGO improves MMLU over PCGrad for every base objective: +1.1 (GA+GD), +0.6 (NPO+GD), +1.0 (SimNPO+GD). Cyber likewise exhibits consistent retention gains over PCGrad (+1.2, +1.0, +0.4). Forget performance is also favorable compared to PCGrad, with only an exception on WMDP Bio GA+GD, where the small increase is negligible given the large retention gain.
On RWKU, the advantage widens: Neighbor retention (All) improves by +13.3 (GA+GD), +13.3 (NPO+GD), and +3.4 (SimNPO+GD) over PCGrad.
These results align with our design goal: retain gradients are never opposed, yielding higher directional fidelity.

\paragraph{(3) Trade-off frontiers shift outward with SAGO.} As shown by the Pareto fronts in Figures~\ref{fig:wmdp_tradeoff} (WMDP) and Figures~\ref{fig:rwku_tradeoff} (RWKU), SAGO expands the performance envelope: at comparable forgetting levels (\textit{e.g.}, Bio forget \(\approx 28\)), SAGO achieves substantially higher retention. On RWKU, SAGO strictly dominates prior points, defining a new frontier in the retention--forgetting trade-off.

\paragraph{Ablation perspective.} The gap between PCGrad and \mname isolates the contribution of sign-aligned gating beyond orthogonal projection.
PCGrad mitigates gradient conflicts through orthogonal projection but may also weaken gradient components that support retention because they are entangled with those conflicts. \mname preserves these useful directions while ensuring no parameter update opposes the retain gradient direction, empirically yielding superior retention performance.

\subsection{Comparison with Other Conflict Mitigation LLM Unlearning Methods}

\begin{table}[t]
    \centering
    \resizebox{\columnwidth}{!}{
    \begin{tabular}{lcc}
    \toprule
    \textbf{Method} & \textbf{Forget $\downarrow$} & \textbf{MMLU $\uparrow$} \\
    \midrule
    \multicolumn{3}{c}{\textbf{WMDP Bio}} \\
    \midrule
    NPO + GD + GRU    & 26.2 & 42.1 \\
    RMU + BLUR       & 27.6 & 53.5 \\
    GA + GD + Global PCGrad & 24.8 & 51.0 \\
    \rowcolor{methodbg} GA + GD + PCGrad & \textbf{24.5} & 53.0 \\
    \rowcolor{methodbg} GA + GD + \mname & 26.0 & \textbf{54.1} \\
    \midrule
    \multicolumn{3}{c}{\textbf{WMDP Cyber}} \\
    \midrule
    NPO + GD + GRU    & \textbf{25.2} & 55.6 \\
    RMU + BLUR       & 25.6 & 55.6 \\
    GA + GD + Global PCGrad & 28.1 & 58.1 \\
    \rowcolor{methodbg} GA + GD + PCGrad & 27.0 & 58.5 \\
    \rowcolor{methodbg} GA + GD + \mname & 26.2 & \textbf{59.7} \\
    \bottomrule
    \end{tabular}
    }
    \caption{\textbf{Comparison with other conflict mitigation methods on WMDP.} Lower forget accuracy indicates better forgetting effectiveness, and higher MMLU reflects better retention. PCGrad denotes our module-wise variant.}
    \label{tab:conflict_comparison}
\end{table}

We compare our method with two conflict-mitigation approaches for LLM unlearning: GRU \citep{wang2025gru} and BLUR \citep{reisizadeh2025blur}. For each baseline, we report the best-performing variant identified in the respective papers. We also include Global PCGrad, which performs projection in the joint parameter space (as in GRU), to contrast with our module-wise PCGrad that projects per module.
As shown in Table~\ref{tab:conflict_comparison}, our module-wise PCGrad already outperforms Global PCGrad, confirming that finer-grained projection better mitigates inter-task interference. \mname further improves retention while maintaining competitive forgetting effectiveness.

\subsection{Analysis of Gradient Geometry}

\begin{table}[t]
    \centering
    \resizebox{0.49\textwidth}{!}{
    \begin{tabular}{lccc}
    \toprule
    \textbf{Method} & \textbf{Forget-Retain} & \textbf{Comb-Forget} & \textbf{Comb-Retain} \\
    \midrule
    GradDiff & $-0.40$ & $\mathbf{0.50}$ & $0.42$ \\
    PCGrad   & $-0.52$ & $0.29$ & $0.52$ \\
    \mname     & $-0.35$ & $0.17$ & $\mathbf{0.57}$ \\
    \bottomrule
    \end{tabular}
    }
    \caption{\textbf{Average cosine similarity among task and synthesized gradients.} Higher Comb-Retain and moderately positive Comb-Forget are desirable for retention-prioritized unlearning.}
    \label{tab:cosine_stats}
\end{table}

To better understand how different synthesis strategies reshape optimization dynamics, we tracked cosine similarities between the raw task gradients (forget and retain) and the final combined gradient (``Comb''). We report the average over 100 training steps on WMDP Cyber in Table~\ref{tab:cosine_stats}.
\textbf{The Forget-Retain similarity} is negative for all methods, which means the two raw tasks naturally pull the model in opposite directions. This conflict is weakest for \mname, suggesting that \mname reshapes the parameter space to reduce conflict between forget and retain gradients.
\textbf{Considering how the final gradient} aligns with the retain gradient (Comb-Retain), \mname yields the greatest similarity. This matches our goal: never move in a direction that goes against retention.
GradDiff attains the largest Comb-Forget similarity, but at the cost of a clear drop in Comb-Retain, meaning the forgetting signal dominates and degrades retention. PCGrad reduces Comb-Forget by projecting away conflicting components. \mname goes further: it keeps only gradient components whose signs already agree with the retain task, yielding (i) the strongest alignment with retention (highest Comb-Retain), and (ii) a controlled, non-excessive contribution from forgetting (moderate Comb-Forget).

\section{Related Work}

\paragraph{LLM Unlearning.}
The rise of large language models has raised significant concerns about safety risks and privacy leaks, increasing interest in methods for LLM unlearning \citep{yao2024large}.
LLM unlearning has a wide range of practical applications, including the protecting copyrighted materials and removing sensitive personal information.
A variety of methods for LLM unlearning have been proposed, ranging from model optimization-based \citep{zhang2024negative,fan2024simplicity,wang2025gru,sondej2025robust} methods to inference-time methods \citep{pawelczyk2024context,suriyakumar2025ucd,ji2024reversing}.
Despite these advances, unlearning in large language models remains challenging because it requires balancing two competing objectives: removing targeted information while preserving the model's overall capabilities \citep{maini2024tofu,shi2024muse,jin2024rwku,li2024wmdp}.
Efforts to address this challenge fall into three categories: (1) objectives that transform the unbounded GA objective into a bounded form to prevent excessive forgetting, uch as NPO \citep{zhang2024negative} and SimNPO \citep{fan2024simplicity}; (2) approaches like GradDiff \citep{liu2022continual} that incorporate an explicit retain objective; and (3) inference-time unlearning methods that modify model outputs without altering model weights \citep{pawelczyk2024context,suriyakumar2025ucd,ji2024reversing}.
Currently, GRU \citep{wang2025gru} uses PCGrad \citep{yu2020gradient} to resolve gradient conflicts between forget and retain gradients. Our work builds on similar ideas but focuses on the asymmetric nature of the two tasks.

\paragraph{Conflict Mitigation in Multi-task Learning.}
Multi-task learning (MTL) refers to learning a single model that can tackle multiple different tasks.
However, learning multiple tasks simultaneously can be a challenging optimization problem.
The most common MTL objective in practice is the weighted loss over all tasks. But directly optimizing the weighted loss is known to lead to undesirable performance.
A known cause of this phenomenon is the conflicting gradients between different tasks \citep{yu2020gradient}.
To address this problem, previous works have proposed various gradient manipulation techniques \citep{yu2020gradient,liu2021conflict,chen2020just,pmlr-v162-navon22a}.
For example, PCGrad \citep{yu2020gradient} seeks a better update vector by projecting one task's gradient onto the normal plane of another task's gradient.
Another line of work is multi-task learning via model merging \citep{yang2024model}. Instead of manipulating gradients, these methods first train separate models for each task and then merge the models into a single one. To mitigate the conflict between different tasks, a variety of methods have been proposed to resolve conflicts among task vectors \citep{yadav2023ties,gargiulo2025task,yu2024language,marczak2025no,sun2025cat}.

\section{Conclusion}

In this paper, we revisited LLM unlearning through an \emph{asymmetric two-task} perspective.
Building on this perspective, we introduced a modular framework for conflict-aware gradient synthesis and instantiated it with (i) a per-parameter adaptation of PCGrad and (ii) \mname, a novel gradient synthesis method that gates forget and retain signals at element level to guarantee non-antagonistic updates. Across WMDP and RWKU, \mname consistently shifts the Pareto frontier outward, recovering substantially more retention at comparable forgetting strength, validating that retention-prioritized gradient geometry matters.

\section*{Limitations}
Our study, while demonstrating consistent retention gains under many settings, still has several limitations. (1) Benchmark scope: We focus on two representative LLM unlearning benchmarks (WMDP and RWKU); broader domains (\textit{e.g.}, multimodal or code models) are left for future exploration. (2) Computational cost: Computing and storing separate retain and forget gradients introduces additional overhead versus a single objective, though still practical in standard finetuning regimes; we do not optimize for extremely low-resource settings.

\bibliography{custom}

\appendix

\section{Hyperparameters} \label{sec:hyperparameters}
Since the evaluation of unlearning involves two opposing metrics---forgetting and retention---there is an inherent trade-off between them. Our hyperparameter tuning protocol is therefore to first match the forgetting metrics across methods as closely as possible, and then compare the methods under this constraint. Concretely, we mainly tune the following hyperparameters:
\begin{itemize}
    \item Weight $\gamma$ of the forgetting objective: the default value is 1.0, and we sweep $\gamma \in [0.1, 1.0]$ when necessary to match the baseline's forgetting performance.
    \item Weight $\alpha$ of the retention objective: the default value is 1.0, and we sweep $\alpha \in [0.1, 1.0]$ when necessary to match the baseline's forgetting performance.
    \item Training budget and learning rate: for WMDP, the default is 100 training steps; for RWKU, we fix the budget to 2 epochs and tune the learning rate and method-specific hyperparameters (\textit{e.g.}, $\beta$ for NPO).
\end{itemize}

\end{document}